\documentclass{ecai} 
  \usepackage{cancel}

\usepackage{latexsym}
\usepackage{amssymb}
\usepackage{amsmath}
\usepackage{amsthm}
\usepackage{booktabs}
\usepackage{enumitem}
\usepackage{graphicx}
\usepackage{color}

\usepackage{colortbl}
\usepackage{cleveref}
\usepackage{bbm}
\usepackage{multirow}

\newtheorem{theorem}{Theorem}

\newcommand{\BibTeX}{B\kern-.05em{\sc i\kern-.025em b}\kern-.08em\TeX}

\begin{document}

%%%%%%%%%%%%%%%%%%%%%%%%%%%%%%%%%%%%%%%%%%%%%%%%%%%%%%%%%%%%%%%%%%%%%%%%

\begin{frontmatter}

%%% Use this command to specify your submission number.
%%% In doubleblind mode, it will be printed on the first page.

%\paperid{6573} 

%%% Use this command to specify the title of your paper.

\title{A Distance Metric for Mixed Integer Programming Instances}

%%% Use this combinations of commands to specify all authors of your 
%%% paper. Use \fnms{} and \snm{} to indicate everyone's first names 
%%% and surname. This will help the publisher with indexing the 
%%% proceedings. Please use a reasonable approximation in case your 
%%% name does not neatly split into "first names" and "surname".
%%% Specifying your ORCID digital identifier is optional. 
%%% Use the \thanks{} command to indicate one or more corresponding 
%%% authors and their email address(es). If so desired, you can specify
%%% author contributions using the \footnote{} command.

%\author{Anonymous authors}

\author[A]{\fnms{Gwen}~\snm{Maudet}\orcid{0000-0003-0340-2542}\thanks{Corresponding Author.  Email: gwen.maudet@uni.lu}}
\author[B]{\fnms{Grégoire}~\snm{Danoy}\orcid{0000-0001-9419-4210}\thanks{Corresponding Author.  Email: gregoire.danoy@uni.lu}}
\address[A]{SnT, University of Luxembourg, Esch-sur-Alzette, Luxembourg}
\address[B]{FSTM/DCS, SnT, University of Luxembourg, Esch-sur-Alzette, Luxembourg}

%%% Use this environment to include an abstract of your paper.

\begin{abstract}
Mixed-integer linear programming (MILP) is a powerful tool for addressing a wide range of real-world problems, but it lacks a clear structure for comparing instances. A reliable similarity metric could establish meaningful relationships between instances, enabling more effective evaluation of instance set heterogeneity and providing better guidance to solvers, particularly when machine learning is involved. Existing similarity metrics often lack precision in identifying instance classes or rely heavily on labeled data, which limits their applicability and generalization. To bridge this gap, this paper introduces the first mathematical distance metric for MILP instances, derived directly from their mathematical formulations. By discretizing right-hand sides, weights, and variables into classes, the proposed metric draws inspiration from the Earth mover's distance to quantify mismatches in weight-variable distributions for constraint comparisons. This approach naturally extends to enable instance-level comparisons. We evaluate both an exact and a greedy variant of our metric under various parameter settings, using the StrIPLIB dataset. Results show that all components of the metric contribute to class identification, and that the greedy version achieves accuracy nearly identical to the exact formulation while being nearly 200 times faster. Compared to state-of-the-art baselines—including feature-based, image-based, and neural network models—our unsupervised method consistently outperforms all non-learned approaches and rivals the performance of a supervised classifier on class and subclass grouping tasks.
\end{abstract}

\end{frontmatter}

%%%%%%%%%%%%%%%%%%%%%%%%%%%%%%%%%%%%%%%%%%%%%%%%%%%%%%%%%%%%%%%%%%%%%%%%

\section{Introduction}
Mixed-integer linear programming (MILP) is a fundamental tool for formulating and solving optimization problems. It involves optimizing an objective function subject to constraints that include both continuous and discrete decision variables. This flexibility allows MILP to address a wide variety of real-world problems across diverse fields, including transportation~\cite{archetti_optimization_2022}, manufacturing~\cite{tiwari_survey_2015}, and e-health services~\cite{shafaghsorkh_application_2022}.

The MILP space remains vast and unstructured in terms of the links between instances, and the introduction of structure within it, as discussed in this paper, yields two primary benefits. First, it facilitates the characterization of sets of instances. For instance, it enables the creation of heterogeneous benchmarks, as exemplified by the MIPLIB 2017 library~\cite{gleixner_miplib_2021}. This also allows for the evaluation of the heterogeneity of an evaluation instance set, offering insights into the generalizability of the assessed methods.  Second, incorporating structural information can significantly enhance solver performance. In particular, machine learning (ML)-based approaches integrated into MILP solvers—commonly referred to as ML-MILPs—have demonstrated efficacy only within restricted subsets of similar instances~\cite{lodi_learning_2017,bengio_machine_2021,cappart_combinatorial_2023,zhang_survey_2023,scavuzzo_machine_2024}. Structuring the MILP space into clusters of homogeneous instances is therefore a critical step toward enabling the broader applicability of ML-MILP techniques across diverse problem domains.

Early attempts to structure the MILP space focused on problem classes definition based on recurring real-world problem types to structure MILPs~\cite{bastubbe_striplib_2025}. However, these classes represent only a subset of the MILP space. To capture the full range of instances, tagging systems have been introduced~\cite{gleixner_miplib_2021}, which assign multiple hierarchical tags to instances, from the most specific to the most general. However, this system does not create disjoint classes, which are necessary for a clean and meaningful structure. To overcome these limitations, a similarity metric between instances would allow for a formal characterization of their relationships, enabling the development of a comprehensive classification of the MILP space. Current similarity methods include the Image-Based Structural Similarity (ISS) approach~\cite{steever_image-based_2022}, which treats constraint matrices as grayscale images analyzed by autoencoders. However, this method is sensitive to the ordering of constraints. Graph Neural Networks (GNNs)~\cite{steever_graph-based_2024} address this by using invariant graph representations, but their applicability is limited to pre-trained problem classes, restricting their generalizability. The MIPLIB 2017 framework~\cite{gleixner_miplib_2021} offers a different approach by using over 100 handcrafted features in a high-dimensional feature space. However, it lacks a rigorous theoretical foundation for feature selection and normalization, which diminishes its robustness. A more comprehensive discussion of these methods can be found in~\cref{related_works}.

This paper presents a training-free mathematical distance for comparing MILP instances, designed to identify instances belonging to the same class as similar solely from their formulation. Variables, weights, and right-hand sides are categorized into discrete classes, allowing the definition of constraint distances based on mismatches in weight-variable pairs and right-hand sides. The overall MILP distance is then computed by matching constraints in a way that minimizes their distance,  combined with the distance between objective functions. To accommodate instances of varying sizes, the method employs a normalized representation that captures the proportions of similar weight-variable pairs and constraints. By demonstrating that this metric aligns with the Earth Mover's Distance (EMD)~\cite{rubner_earth_2000}, we establish that it satisfies the properties of a mathematical distance. For practical use, we introduce a greedy heuristic for efficient metric computation, ensuring seamless integrations with reduced computational cost. This is detailed in~\cref{distance_def}.

We conduct a series of evaluations using the StrIPLIB library~\cite{bastubbe_striplib_2025}, which hierarchically organizes MILP instances into well-defined classes and subclasses. Two experiments are performed: the first replicates the protocol of~\cite{steever_graph-based_2024} to assess the ability to identify instances from the same class (19 in total), while the second focuses on three selected classes to evaluate subclass-level grouping. In both cases, we measure performance by computing the proportion of nearest neighbors that belong to the same class (or subclass) as each test instance.
We evaluate several variants of our distance metric, each omitting a specific component (e.g., weights, variables, right-hand sides, or objective function) to assess its contribution. Results show that each component plays a meaningful role in improving class identification. We also compare our greedy version with the exact one, observing a performance gap under $4\%$ for 17 of the 19 classes, while achieving a 200× speedup.
Compared to state-of-the-art similarity metrics~\cite{steever_image-based_2022, steever_graph-based_2024, gleixner_miplib_2021}, our method outperforms both the feature-based~\cite{gleixner_miplib_2021} and image-based~\cite{steever_image-based_2022} approaches, and matches the performance of the GNN-based method~\cite{steever_graph-based_2024}, despite requiring no supervised training. Full experimental details are provided in~\cref{evaluation}.

Our contributions can be listed as follows:
\begin{itemize}
    \item Identification of the challenge in structuring the MILP space, particularly pertinent in the context of ML, which can be addressed through a similarity metric.
    \item Introduction of a normalized representation of MILP instances, leveraging feature classification to enable meaningful comparisons across instances of varying sizes.
    \item Definition of a mathematical distance that captures structural patterns within and across constraints.
    \item Comprehensive experiments with new benchmarks and additional baselines, demonstrating our approach’s efficiency.
\end{itemize}

\section{Related Works and Positioning}~\label{related_works}
In this section, we present the latest advancements in ML applied to MILP solvers, highlighting the current limitations with regard to generalizability. Next, we discuss the existing efforts for structuring the MILP space, starting with instance categorization methods and followed by similarity approaches between instances. Finally, we explain how our approach addresses the shortcomings of existing methods. We emphasize that our method has the potential to resolve the generalizability issue faced by ML-MILPs.

\subsection{ML for MILP}

ML has played a growing role in advancing MILP solvers by automating key components such as branching strategies, search heuristics, and cutting plane selection, leading to substantial performance improvements~\cite{lodi_learning_2017,bengio_machine_2021,cappart_combinatorial_2023,scavuzzo_machine_2024,zhang_survey_2023}. State-of-the-art on ML-MILPs, including those proposed in~\cite{he_learning_2014,gasse_exact_2019,labassi_learning_2022,tang_reinforcement_2020,nair_learning_2018,thuerck_learning_2023,paulus_learning_2023}, are typically trained on synthetic datasets or well-established benchmarks limited to specific problem classes. Their generalization is usually evaluated by scaling instance size, rather than by assessing structural diversity. As a result, these models often fail to generalize effectively to heterogeneous or real-world instances. Although heterogeneous benchmarks such as MIPLIB 2017~\cite{gleixner_miplib_2021} have been employed to assess the performance of ML-enhanced techniques—such as cutting plane generation or adaptive search policies~\cite{turner_adaptive_2023,maudet_search_2024}—results have often been mixed due to challenges related to model convergence and the selection of suitable training sets.

At present, ML-MILP models generally require structurally homogeneous instance sets—i.e., instances drawn from the same underlying problem type, albeit with varying sizes. These models continue to struggle with generalization across the entire MILP space. A promising direction to address this limitation is to partition the MILP space into groups of structurally similar instances. Such structured partitioning would enable the design of specialized ML-MILP models tailored to each group, thereby improving generalizability.

Furthermore, most existing studies define their own evaluation sets, often without a principled assessment of heterogeneity. This practice complicates the comparison of different ML-MILP approaches in terms of their generalization capacity. Establishing a rigorous characterization of heterogeneity within evaluation datasets is thus essential for meaningful benchmarking and fair performance comparisons.

On these two grounds, we review existing approaches from two complementary angles: (i) methods for partitioning the MILP space and (ii) techniques for evaluating heterogeneity between instances.

\subsection{Categorization of MILP Instances}

Efforts to partition the MILP instance space have a long history. Early work focused on categorizing well-known problem domains in logistics and flow management~\cite{ausiello_structure_1977,papadimitriou_optimization_1988}, which facilitated the development of specialized solution techniques targeted at specific problem types. More recently, StrIPLIB~\cite{bastubbe_striplib_2025} has extended this by proposing a dataset of over 21,000 MILP instances drawn from the literature and organizing them into 33 hierarchically structured classes and subclasses.
While this represents a significant step toward formalizing the MILP instance space, the classification framework exhibits several shortcomings. Many classes are not mutually exclusive; for example, the General Assignment problem reduces to a Knapsack problem when restricted to a single agent. Furthermore, the scope and granularity of the classes vary widely. Some categories, such as Bin Packing, encompass thousands of instances (e.g., 3,640), whereas others—like Binary/Ternary Code Construction—contain only a handful (e.g., 2). As a result, certain classes disproportionately dominate the dataset, while others cover only narrow and specialized niches, raising concerns about the structural balance and homogeneity of this existing classification appraoch.
Additionally, this categorization is far from exhaustive. Real-world MILP problems often fall outside the boundaries of the predefined classes, rendering the current classification schemes insufficient for capturing the full diversity of the MILP space.

In parallel, MIPLIB 2017~\cite{gleixner_miplib_2021}, a benchmark library of real-world instances, employs a constraint tagging system to describe instance structures. These tags are linked to manually defined constraint templates, allowing for the characterization of instances directly from their formulations.  While this approach enables broad coverage—any instance can be described using a combination of tags—the system does not support partitioning. Tags are inherently hierarchical, ranging from general to highly specific, and instances often receive multiple tags, precluding a clean division of the space into disjoint, homogeneous groups.

Despite these initiatives, a robust and comprehensive segmentation of the MILP space into disjoint, structurally homogeneous classes remains an open challenge. Achieving such a segmentation would likely require the development of a principled and reliable similarity metric capable of quantifying structural relationships between instances.

\subsection{Similarity Measures Between MILPs}
Various methods have been proposed to assess the similarity between MILP instances, either to guide solvers in the resolution process or to evaluate the heterogeneity of instance sets. For the first objective, Steever \emph{et al.} explored similarity methods, emphasizing that the structure of a MILP can reveal valuable insights into its solution process without more details. Their initial approach, the ISS method~\cite{steever_image-based_2022}, represents instances by encoding constraint matrices as grayscale images analyzed via autoencoders, producing a lower-dimensional representation of the input instance. Subsequently, they introduced a GNN-based approach~\cite{steever_graph-based_2024}, where an input graph representation of an instance is processed to output the probability of the instance belonging to each of the trained classes, using training data. The second objective, evaluating heterogeneity, was pursued in the construction of the MIPLIB2017 library~\cite{gleixner_miplib_2021}, where their similarity metric was employed to define instance libraries that maximize heterogeneity. This metric represents instances through 100 handcrafted features, including attributes such as problem size, coefficient distributions, and categorization tags. In both methods, the similarity between instances is measured using the Euclidean distance in the output space.

Despite these contributions, existing similarity measures face important limitations. Unsupervised approaches such as the feature-based method lack rigorous theoretical justification for feature selection, which limits confidence in their applicability across diverse MILP formulations. The ISS method has been shown to underperform compared to the GNN-based alternative. On the other hand, the supervised GNN model relies on predefined instance classes during training—a framework that lacks theoretical coherence for defining a global, principled structure over the MILP space. This reliance introduces significant difficulties in selecting appropriate training sets, especially when aiming to generalize to the full diversity of MILP problems.

In summary, existing similarity metrics remain constrained either by poor performance for unsupervised appraoch or by the need for supervised training on ill-defined class labels, motivating the search for alternative, mathematically grounded approaches that can support unsupervised structuring of the MILP space.

\subsection{Positioning}
To overcome the limitations identified in previous approaches, we propose a lightweight similarity metric for MILP instances that is directly derived from their mathematical formulations and supported by well-defined mathematical properties. Instead of relying on predefined, discrete problem classes, our approach generalizes this concept by defining similarity as a continuous measure of structural closeness. Within a given problem class—assuming a consistent representation—strong structural patterns tend to persist across instances, even as instance sizes vary. That is, larger instances may introduce additional variables or constraints, yet the fundamental structure (e.g., variable types, constraint templates, and coefficient distributions) typically remains stable. The goal is thus to identify these persistent structural similarities that are preserved as instance size increases.

The proposed metric not only facilitates the selection of evaluation sets by controlling heterogeneity—similar to the motivation in~\cite{gleixner_miplib_2021}—but also enables the quantitative assessment of heterogeneity in existing benchmarks. This, in turn, provides a principled means for evaluating the generalizability of optimization and learning methods. Moreover, as noted by Steever \emph{et al.}, such similarity metrics can be leveraged to guide solver behavior, enhancing performance. From an ML perspective, a reliable similarity metric allows for the segmentation of the MILP space into structurally coherent subsets. This enables training ML-MILP models on homogeneous instance groups, which improves convergence and mitigates the difficulties associated with training set selection. In contrast to handcrafted heuristics—often tailored to specific problem families—ML methods are inherently data-driven and adapt to the structure of the training data. As such, defining similarity directly from instance formulations, rather than relying on problem-type semantics, presents a promising approach for developing ML-MILP methods that are both scalable and broadly applicable. 

Nevertheless, the task of explicitly partitioning the MILP space—though a natural extension of our metric—is beyond the scope of this paper. Our primary goal here is to validate the relevance and robustness of the proposed distance measure itself.

%which rely on prior domain knowledge, similarity-based segmentation offers a general framework for learning across instance groups defined purely through structural similarity.

%where \( C = (c_i)_{1 \leq i \leq m_\mathbf{P}} \) is the set of constraints with \( b_i \in \mathbb{R} \); \( V_i = (v_i^j)_{1 \leq j \leq n_i} \) are the non-zero variables in constraint \( c_i \); \( W_i = (w_i^j)_{1 \leq j \leq n_i} \) are the weights associated with \( c_i \), with \( w_i^j \in \mathbb{R}^* \); and the objective function's weights and variables are \( (w_z^j)_{1 \leq j \leq n_z} \) and \( (v_z^j)_{1 \leq j \leq n_z} \), respectively.

\section{Formal Definition of the Distance}~\label{distance_def}

In this section, we formally define a distance metric for comparing MILP instances, with the goal of identifying structurally similar instances—even when their sizes differ—by leveraging common patterns typically observed within instances of the same problem class. The metric is constructed by categorizing variables, their corresponding weights, and right-hand sides, followed by introducing a normalized instance representation aimed at eliminating dimension-dependent parameters. Using this representation, we define a distance between individual constraints and extend it to entire MILP instances by comparing proportions of similar weight-variable pairs and constraints, rather than direct pairwise comparisons. This metric satisfies the properties of a mathematical distance. having similarities with the well known EMD. 

\subsection{Background}

A MILP instance \( \mathbf{P} \) can be expressed mathematically as~\eqref{MILP_def}:

\begin{equation}~\label{MILP_def}
    \begin{aligned}
    & \text{Minimize } z = \sum_{j \in \mathbb{N}_0} w_0^j v^j, \\
    & \text{Subject to}\quad  i \in \{1, \dots, m\},\;  c_i: \sum_{j \in \mathbb{N}_i} w_i^j v^j \leq b_i, \; 
\end{aligned}
\end{equation}

where the key components are defined as follows:
\begin{itemize}
    \item \( (v^j)_{1 \leq j \leq n} \): the set of variables in the instance.
    \item \( C = (c_i)_{1 \leq i \leq m} \): the set of constraints, with \( c_0 \) conventionally denoting the objective function. For each constraint \( c_i \), \( \mathbb{N}_i \subseteq \{1, \dots, n\} \) represents the indices of variables with non-zero weights, with \( n_i = |\mathbb{N}_i| \) the number of variables in the constraint.
    \item \( b_i \in \mathbb{R} \): the right-hand side of constraint \( c_i \).
    \item \( (v^j)_{j \in \mathbb{N}_i} \): the variables associated with constraint \( c_i \).
    \item \( (w_i^j)_{j \in \mathbb{N}_i} \): the weights associated to the variables in \( c_i \).
\end{itemize}

For clarity, when necessary, we denote \( c_{i}(\mathbf{P}) \) to explicitly refer to constraint (or objective function) \( c_i \) associated with instance \( \mathbf{P} \).

\subsection{Classification of MILP Features}
To define the distance between constraints—and by extension, entire MILP instances—we classify the three essential features of MILP formulations: right-hand sides, variables, and weights. This classification discretizes these features into distinct classes, enabling simplified comparisons through binary similarity measures, while also accounting for redundancies both within and across constraints.

\subsubsection*{Formalism of MILP Feature Classes}
The classification process assigns classes to elements as follows:  
\begin{itemize}
    \item Each right-hand side \( b_i \) is assigned a class \( t(b_i) \in \mathcal{C}(b) \).
    \item Each variable \( v^j \) is assigned a class \( t(v^j) \in \mathcal{C}(v) \).
    \item Each weight \( w_i^j \) is assigned a class \( t(w_i^j) \in \mathcal{C}(w)\).
\end{itemize}  

To define the distance, we use the indicator function \( \mathbbm{1} \), which compares two elements \( x_1 \) and \( x_2 \) of type \( x \) (right-hand sides, variables, or weights). If \( x_1 \) and \( x_2 \) are in the same class, \( \mathbbm{1}(t(x_1) \neq t(x_2)) = 0 \); otherwise, it equals 1.

\subsubsection*{Choices Made for the Classification}
For variables, we use the standard classification into three categories: binary (\( B \)): $v^j\in\{0,1\}$, integer (\( I \)): $v^j\in\mathbb{Z}$, and continuous (\( C \)): : $v^j\in\mathbb{R}$. Constraints on lower and upper bounds are excluded to maintain a reduced number of classes.
For weights and right-hand sides, the domains are \( \mathbb{R} \) and \( \mathbb{R}^* \), respectively. To define classes, instead of defining intervals, which may introduce boundary challenges, we classify by isolating the most frequently occurring singletons as distinct classes. The remaining values form a complementary class encompassing all other elements.
The classification is derived from the MIPLIB 2017 collection, a comprehensive dataset of 1065 MILP instances~\cite{gleixner_miplib_2021} representing diverse MILP problems. Analysis of this dataset identifies \( -1 \) (40\%) and \( 1 \) (33\%) as the most frequent singletons for weights, and \( 0 \) (57\%) and \( 1 \) (21\%) for right-hand sides, with high occurrence rates validating the singleton-based approach. Similar patterns are observed in the strIPLIB instances used for evaluation.
In summary, the classification scheme is as follows:
\begin{itemize}
    \item  \( \mathcal{C}(v) = \{B, I, C\} \) for variables, 
    \item \( \mathcal{C}(w) = \{-1, 1, \mathbb{R} \setminus \{-1, 1\}\} \) for weights, 
    \item \( \mathcal{C}(b) = \{0, 1, \mathbb{R} \setminus \{0, 1\}\} \) for right-hand sides. 
\end{itemize}
This discretization scheme is motivated by principles from entropy-based discretization~\cite{witten_data_2011}, aiming to preserve the most informative distinctions in the data. Additionally, by using the same number of classes across different features and ensuring balanced class proportions, we improve comparability of feature importances and prevent any single feature type from dominating the similarity computation.

\subsection{Normalized Representation of an Instance}
From this classification, redundancies are identified both within constraints, in terms of repeated weight-variable pairs, and across constraints with identical constraint representations. To capture repetitions of weight-variable pairs, let \( n(\hat{w}_j, \hat{v}_k, i) \) represent the number of occurrences of the pair \( \hat{w}_j, \hat{v}_k \) in constraint \( c_i \), and define their proportion as \( p_c(\hat{w}_j, \hat{v}_k, i) = \frac{n(\hat{w}_j, \hat{v}_k, i)}{n_i} \). Similarly, redundancies between structurally identical constraints are addressed by identifying the set of structurally unique constraints \( (c_i)_{i \in \mathbb{M}(\mathbf{P})} \), where \( \mathbb{M}(\mathbf{P}) \subseteq \{1, \dots, m\} \), and for \( i \in \mathbb{M}(\mathbf{P})  \), \( m_i \) denotes the number of repetitions of constraint \( c_i \). The proportion of a constraint \( c_i \) within the instance \( \mathbf{P} \) is given by \( p_{\mathbf{P}}(\mathbf{P},c_i) = \frac{m_i}{m} \). Thus, we can define a normalized representation of an instance by removing the dimensions induced by the instance itself, representing the proportions of occurrence of weight-variable pairs within a constraint and the proportions of occurrence of constraints within the instance. Using the notation $\sum_{\hat{w}_j, \hat{v}_k} = \sum_{\hat{w}_j\in \mathcal{C}(w)} \sum_{\hat{v}_k\in  \mathcal{C}(v)}$, we define the normalized version of an instance \( \mathbf{P} \) as in~\eqref{MILP_compact}:

\begin{equation}~\label{MILP_compact}
    \begin{aligned}
    & \text{Minimize } z = \sum_{\hat{w}_j, \hat{v}_k} p_c(\hat{w}_j, \hat{v}_k, 0) \hat{w}_j \hat{v}_k, \\\\
    &\text{Subject to} \quad  i \in \mathbb{M}(\mathbf{P}),\\
    &  p_{\mathbf{P}}(\mathbf{P},i) \times c_i: \sum_{\hat{w}_j, \hat{v}_k} p_c(\hat{w}_j, \hat{v}_k, i) \hat{w}_j \hat{v}_k \leq t(b_i).
\end{aligned}
\end{equation}

\begin{table}[htbp]
\caption{\textbf{Normalized representation of the \emph{app1-2} instance.}  
The first row displays the objective function to minimize, followed by the list of constraints. The first column represents the proportional occurrence (denoted as e$x=10^x$) of each constraint type, and the second column shows the constraint representation.  
In each constraint, a weight-variable pair is denoted as \emph{the proportion of that pair $\times$ the weight class $\cdot$ the variable class}. The class $\mathbb{R}$ represents the class excluding singletons for both the weight and the right-hand side.
}
\begin{tabular}{@{}c|l@{}}
\multicolumn{2}{c}{Minimize $1.0 \times -1 \cdot \text{B}  $}\\\hline
\textbf{Prop} & \textbf{Constraint Representation}  \\ \hline
 $2.0\text{e-$1$}$ & $0.33\times \mathbb{R} \cdot \text{B} +0.33\times 1 \cdot \text{C} +0.33\times -1 \cdot \text{C} \leq \mathbb{R} $\\
 $2.0\text{e-$1$}$ & $0.33\times 1 \cdot \text{B} +0.33\times -1 \cdot \text{C} +0.33\times 1 \cdot \text{C} \leq 1 $\\
  $2.0\text{e-$1$}$ & $0.5\times 1 \cdot \text{C} +0.5\times -1 \cdot \text{C} \leq 0 $\\
  $1.9\text{e-$1$}$ & $0.83\times \mathbb{R} \cdot \text{C} +0.17\times -1 \cdot \text{C} \leq 0 $\\
 $1.9\text{e-$1$}$ & $0.83\times \mathbb{R} \cdot \text{C} +0.17\times 1 \cdot \text{C} \leq 0 $\\
 $4.5\text{e-$3$}$ & $0.8\times \mathbb{R} \cdot \text{C} +0.2\times -1 \cdot \text{C} \leq 0 $\\
 $4.5\text{e-$3$}$ & $0.8\times \mathbb{R} \cdot \text{C} +0.2\times 1 \cdot \text{C} \leq 0 $\\
  $4.0\text{e-$3$}$ & $1.0\times -1 \cdot \text{B} \leq \mathbb{R} $\\
 $2.1\text{e-$3$}$ & $0.33\times \mathbb{R} \cdot \text{B} +0.33\times -1 \cdot \text{C} +0.33\times 1 \cdot \text{C} \leq \mathbb{R} $\\
 $2.1\text{e-$3$}$ & $0.33\times 1 \cdot \text{B} +0.33\times 1 \cdot \text{C} +0.33\times -1 \cdot \text{C} \leq 1 $\\
 $2.1\text{e-$3$}$ & $0.5\times -1 \cdot \text{C} +0.5\times 1 \cdot \text{C} \leq 0 $\\
$1.5\text{e-$5$}$ & $1.0\times -1 \cdot \text{C} \leq \mathbb{R} $\\
 $1.5\text{e-$5$}$ & $1.0\times 1 \cdot \text{C} \leq 1 $\\
\hline
\end{tabular}
\label{app_1_2}
\end{table}

An example is provided in Table~\ref{app_1_2}, where the normalized version of the instance \emph{app1-2} from MIPLIB 2017 is illustrated (\url{https://MIPlib.zib.de/instance_details_app1-2.html}). This instance consists of 53,467 constraints\footnote{Additional constraints were introduced in our model to convert equality constraints into two inequality constraints.} and 26,871 variables, yet only 13 unique constraint representations, each containing at most three distinct types of weight-variable pairs. 

It is worth noting that a strong similarity can be drawn with the MIPLIB 2017 tagging process, which constructs manually defined constraint templates~\footnote{See \url{https://MIPlib.zib.de/statistics.html}}. Our standardization approach enables the definition of these templates while also generalizing the process through the automatic definition of templates that compactly represent the instance.
%The weight classes are $\{-1, 1, \text{else}\}$, the variable classes are $\{B, C, I\}$, and the right-hand side classes are $\{0, 1, \text{else}\}$. 

\subsection{Distance Definitions}
Our objective is to define a distance measure between MILP instances that effectively groups instances from the same problem class, even when their sizes differ. Specifically, by analyzing modifications of similar problem types of varying sizes, we aim to assign zero distance to: 
(i)  constraints that vary in the number of variables but maintain proportionality in the distribution of weight-variable pairs;
(ii) instances with different numbers of constraints but the same proportion of similar constraints.
Our distance measure is based on the normalized representation in~\eqref{MILP_compact}, where we aim to minimize mismatches in the weight-variable pairs for constraint comparison, and minimize the distances between constraints to determine the overall distance between instances.

\subsubsection*{Distance between Weight-Variable Pairs}

First, to compare a weight-variable pair, we establish the following rules: 
(i) if the weights belong to different classes, we add \( \alpha > 0 \);  (ii) if the variables belong to different classes, we add \( \beta > 0 \).
Therefore, for a pair of weight and variable  \( \hat{w}_j, \hat{v}_k \) and another pair \( \hat{w}_l, \hat{v}_o \), their similarity is expressed as~\eqref{distance_qairs}:
\begin{equation}\label{distance_qairs}
    d_{w,v}(\hat{w}_j, \hat{v}_k, \hat{w}_l, \hat{v}_o) = \alpha \mathbbm{1}(\hat{w}_j \neq \hat{w}_l) + \beta \mathbbm{1}(\hat{v}_k\neq \hat{v}_o)
\end{equation}
$d_{w,v}$ represents the discrete distance metric, it satisfies the properties of a mathematical distance.
\begin{theorem}
The function \( d_{w,v} \) defined in \eqref{distance_qairs} is a mathematical distance between elements of \( \mathcal{C}(w)\times\mathcal{C}(v) \).
\end{theorem}

%&\text{Subject to:} \\
\subsubsection*{Distance Between Constraints}  
The distance between two constraints, \( c_q \) and \( c_r \), is defined as the minimal transfer of proportions of weight-variable pairs from \( c_q \) to \( c_r \). This transfer is weighted by \( d_{w,v} \), and it includes an additional term accounting for potential differences in their right-hand sides, weighted by $\gamma$. Considering the proportions of weight-variable pair occurrences \( (p(\hat{w}_j, \hat{v}_k , q))_{\hat{w}_j, \hat{v}_k } \) for \( c_q \) and \( (p(\hat{w}_l, \hat{v}_o , r))_{\hat{w}_l, \hat{v}_o } \) for \( c_r \), the simiarity is expressed mathematically as~\eqref{distance_constraints}:

\begin{equation}~\label{distance_constraints}
\begin{aligned}
&d_c(c_q, c_r) = \\
    &\min \sum_{\hat{w}_j, \hat{v}_k } \sum_{\hat{w}_l, \hat{v}_o} d_{w,v}(\hat{w}_j, \hat{v}_k, \hat{w}_l, \hat{v}_o) \cdot f(\hat{w}_j, \hat{v}_k, \hat{w}_l, \hat{v}_o) \\
    &+ \gamma\mathbbm{1}(t(b_q) \neq t(b_r)),
\end{aligned}
\end{equation}
subject to the following constraints given in~\eqref{distance_constraints_1}:
\begin{equation}~\label{distance_constraints_1}\tag{\ref{distance_constraints}.1}
\begin{aligned}
&\forall \hat{w}_j, \hat{v}_k, \hat{w}_l, \hat{v}_o,f(\hat{w}_j, \hat{v}_k, \hat{w}_l, \hat{v}_o)\geq 0, \\
&\forall \hat{w}_j, \hat{v}_k, \sum_{\hat{w}_l, \hat{v}_o} f(\hat{w}_j, \hat{v}_k, \hat{w}_l, \hat{v}_o) = p_c(\hat{w}_j, \hat{v}_k, q),\quad \\
&\forall \hat{w}_l, \hat{v}_o, \sum_{\hat{w}_j, \hat{v}_k} f(\hat{w}_j, \hat{v}_k, \hat{w}_l, \hat{v}_o) = p_c(\hat{w}_l, \hat{v}_o, r).
\end{aligned}
\end{equation}

Here, \( f(\hat{w}_j, \hat{v}_k, \hat{w}_l, \hat{v}_o) \) represents the transfer of proportions from the weight-variable pair \(\hat{w}_j, \hat{v}_k\) in \( c_q \) to the pair \( \hat{w}_l, \hat{v}_o \) in \( c_r \). The constraints ensure that the total transfer from all pairs \( \hat{w}_j, \hat{v}_k\) equals the proportion \( p_c(\hat{w}_j, \hat{v}_k, q) \), and similarly, the total transfer to \( \hat{w}_l, \hat{v}_o\) equals \( p_c(\hat{w}_l, \hat{v}_o, r) \).  Notably, this corresponds to the EMD~\cite{rubner_earth_2000}, also commonly referred to as the Wasserstein distance~\cite{berger_wasserstein_2009}. This distance quantifies the minimum amount of change required to transition from one distribution to another. Combined with an additional distance component, this metric satisfies the properties of a mathematical distance.

%(in our case, from one distribution of weight-variable pairs to another)

\begin{theorem}
The function \( d_c \) defined in \eqref{distance_constraints} is a mathematical distance between constraints represented as~\eqref{MILP_compact}.% represented as~\eqref{MILP_compact}: \( \{(p_{\hat{w}_j,\hat{v}_k})_{\hat{w}_j\in \mathcal{C}(w), \hat{v}_k\in \mathcal{C}(v) },\sum_pp_{\hat{w}_j,\hat{v}_k}=1\}\times\mathcal{C}(b)\)
\end{theorem}

\subsubsection*{ Distance Between Instances}  
The distance between two instances is defined using the previously established constraint distance. For instances \( \mathbf{P_s} \) and \( \mathbf{P_t} \), the set \( ( c_{i_s}(\mathbf{P_s}),p_{\mathbf{P}}(\mathbf{P_s},c_{i_s}(\mathbf{P_s}))_{i_s \in \mathbb{M}(\mathbf{P_s})} \) represents distinct constraints and their proportions for \( \mathbf{P_s} \), with a similar representation for \( \mathbf{P_t} \). The distance is calculated by transferring the proportions of constraints from \( \mathbf{P_s} \) to \( \mathbf{P_t} \), weighted by \( d_c \), and including an additional term weighted by \( \zeta>0 \) to account for the objective function distance, as shown in~\eqref{distance_instances}:

\begin{equation}~\label{distance_instances}
\begin{aligned}
&d_{\textbf{P}}(\mathbf{P_s}, \mathbf{P_t}) = \\
&\min \sum_{i_s \in \mathbb{M}(\mathbf{P_s})} \sum_{i_t \in \mathbb{M}(\mathbf{P_t})} d_c(c_{i_s}(\mathbf{P_s}), c_{i_t}(\mathbf{P_t})) \cdot f(i_s, i_t) \\\\
&+ \zeta d_c(c_0(\mathbf{P_s}), c_0(\mathbf{P_t})),
\end{aligned}
\end{equation}  

subject to the following constraints given in~\eqref{distance_instances_1}:
\begin{equation}~\label{distance_instances_1}\tag{\ref{distance_instances}.1}
\begin{aligned}
%&\text{Subject to:} \\
&\forall i_s\in \mathbb{M}(\mathbf{P_s}), \forall i_t\in \mathbb{M}(\mathbf{P_t}),f(i_s, i_t) \geq 0,   \\
& \forall i_s\in \mathbb{M}(\mathbf{P_s}),\sum_{i_t \in \mathbb{M}(\mathbf{P_t})} f(i_s, i_t) = p_{\mathbf{P}}(\mathbf{P_s},c_{i_s}(\mathbf{P_s})), \\
&\forall i_t\in \mathbb{M}(\mathbf{P_t}),\sum_{i_s \in \mathbb{M}(\mathbf{P_s})} f(i_s, i_t) = p_{\mathbf{P}}({P_t},c_{i_t}(\mathbf{P_t})).
\end{aligned}
\end{equation}  

Similarly to the distance between constraints, \( f(i_s, i_t) \) represents the transfer of proportions from the constraint \( c_{i_s}(\mathbf{P_s}) \) to \( c_{i_t}(\mathbf{P_t}) \), with constraints ensuring that the transfer respects the proportions \( p_{\mathbf{P}}(\mathbf{P_s}, i_s) \) and \( p_{\mathbf{P}}(\mathbf{P_t}, i_t) \). This first term corresponds to the EMD, ensuring that, when combined with another distance, \( d_{\mathbf{P}} \) adheres to the properties of a mathematical distance.

\begin{theorem}
The function \( d_{\textbf{P}} \) defined in \eqref{distance_instances} is a mathematical distance between instances represented as~\eqref{MILP_compact}.
\end{theorem}  

This metric enables the formal comparison of MILP instances, supports defining neighborhoods of instances, and facilitates classification methods that rely on a valid distance measure.
Our goal was to design a distance metric capable of identifying as similar the instances belonging to the same type, even when their sizes differ.  Specifically, when identical constraints (i.e., constraints with similar proportions of weight-variable pairs) appear in the same proportions across the compared instances, the distance is \( d_{\textbf{P}}=0 \). In more general cases, the proposed distance quantifies the minimum proportion of changes necessary to transform one instance into another,  based on the normalized representation of~\eqref{MILP_compact}. In particular, a change in weight incurs a penalty of \( \alpha \), a change in variables incurs a penalty of \( \beta \), a modification to the right-hand side incurs a penalty of \( \gamma \), and an additional weighting factor \( \zeta \) is applied to account for differences in the objective functions.
Still, a known limitation of this formulation is its flexibility with respect to dimensionality: for example, a constraint involving a single variable can have zero distance from one in which that variable appears multiple times, even though their semantics may differ significantly.

\subsection{Greedy Heuristic for Distance Computation}

The exact computation of the EMD has a complexity of \( O(n^3\log(n)) \)~\cite{orlin_faster_1988}, which presents a significant computational challenge due to its application in both constraint-level and instance-level distance calculations. To address this, we propose an alternative greedy heuristic that iteratively matches the closest pairs, first at the constraint level and subsequently at the instance level, offering a computationally efficient and straightforward approach. At the constraint level, weight-variable pairs from one constraint are iteratively matched with their closest counterparts (based on \( d_{w,v} \)), with matching proportions determined by the smaller of the two proportions. A constraint contains at most \( |\mathcal{C}(w)| \times |\mathcal{C}(v)| \) unique pairs, where \( |\mathcal{C}(\cdot)| \) denotes the number of elements in the set. Thus, comparing two constraints has a complexity of \( O(|\mathcal{C}(w)|^2 \times |\mathcal{C}(v)|^2) \). Extending this to compare all constraints of instances \( \mathbf{P_s} \) and \( \mathbf{P_t} \), the total complexity for constraint-level comparisons becomes \( O(|\mathcal{C}(w)|^2 \times |\mathcal{C}(v)|^2 \times |\mathbb{M}(\mathbf{P_s})| \times |\mathbb{M}(\mathbf{P_t})|) \), where \( |\mathbb{M}(\cdot)| \) represents the number of distinct constraints in an instance. At the instance level, constraints from \( \mathbf{P_s} \) are iteratively matched with their closest counterparts in \( \mathbf{P_t} \) (based on \( d_c \)), again matching proportions according to the smaller value. This step adds a complexity of \( O(|\mathbb{M}(\mathbf{P_s})| \times |\mathbb{M}(\mathbf{P_t})|) \), which remains negligible when compared to the constraint distance computation.

\section{Evaluation}~\label{evaluation}

The main objective of our experimental study is to evaluate the ability of similarity measures to group structurally related MILP instances. 
While problem classification is not without limitations, instances within the same class typically share key structural traits—which is precisely what we aim to capture. We leverage this principle by using the strIPLIB dataset~\cite{bastubbe_striplib_2025}, which contains over 21{,}000 MILP instances organized into hierarchically structured classes and subclasses. We deliberately choose not to use MIPLIB 2017~\cite{gleixner_miplib_2021}, as its real-world instances often span multiple problem types, making class-based evaluation less suitable. We recall that some modeling choices in our distance formulation—such as the categorization of weights and right-hand sides—are derived from empirical observations on MIPLIB instances.

Two experiments are conducted to evaluate the performance of our similarity metric. The first replicates the methodology proposed in~\cite{steever_graph-based_2024}, which assesses the capability to identify instances belonging to the same class. The second extends this analysis by examining the ability to group instances within the same subclass for classes already studied. All source code and resources are publicly available on the Git repository at \url{https://gitlab.com/uniluxembourg/snt/pcog/ultrabo/clustering-for-search-strategy}.

For each experiment, we sample 50 instances per class (or subclass), indexed by \( c \in \mathcal{C} \), where \( \mathcal{C} \) is the set of all classes considered. Each class provides 10 test instances \( \mathbf{P}_c^i(t),i\in[1,10] \) and 40 reference instances \( \mathbf{P}_c^j(r),j\in[1,40] \).
To evaluate a similarity measure (ours or a baseline), we compute distances \( d(\mathbf{P}_c^i(t), \mathbf{P}_{\acute{c}}^j(r)) \) between each test instance and all reference instances across all classes \( \acute{c} \in \mathcal{C} \). For each test instance, we select the 40 reference instances with the smallest distances. Let \( \mathcal{I} = \{ (I_{\acute{c}} \subset [1,40])_{\acute{c} \in \mathcal{C}} \mid \sum_{\acute{c}} |I_{\acute{c}}| = 40 \} \) denote the set of all valid index selections; the optimal combination solves:
\[
\arg \min_{(I_{\acute{c}}) \in \mathcal{I}} \sum_{\acute{c} \in \mathcal{C}} \sum_{j \in I_{\acute{c}}} d(\mathbf{P}_c^i(t), \mathbf{P}_{\acute{c}}^j(r)).
\]

The top-40 accuracy is defined as \( \frac{|I_c|}{40} \), i.e., the proportion of selected reference instances from the correct class. The final score for each class is the average of this value over its 10 test instances. It is important to note that each added class contributes 40 additional comparison points per test instance. As such, performance is influenced by the total number of (sub)classes considered, and results are only comparable within the same experiment, not across experiments.
To enhance clarity, the highest score for a class (or subclass) among baselines is highlighted in green in the result tables. Conversely, if the performance is at least $50\%$ worse than the best score, the corresponding cell is marked in red, with intermediate values represented by a gradient transitioning between these colors.

Our method, denoted \emph{Formal}, depends on parameters \(\alpha\), \(\beta\), \(\gamma\) and \(\zeta\), which respectively control the influence of weights, variables, right-hand sides and the objective function in the distance computation. In the first experiment, we conduct a sensitivity analysis: we test configurations where one parameter is deactivated (denoted with a $\cancel{\cdot}$ symbol), and compare these to the baseline setting where all parameters are set to 1 (denoted $\emptyset$). All sensitivity variants use the greedy version of the distance computation for efficiency. We also report results from the exact version of our metric (denoted $\emptyset E$), and include computation times for both greedy and exact variants. For the second experiment, we restrict our comparison to the greedy version with \(\alpha = \beta = \gamma=\zeta = 1\).

We compare our proposed metric against the three main similarity measures available in the literature—these are, to the best of our knowledge, the only established methods for assessing similarity between MILP instances:
\begin{itemize}
    \item \textbf{Features-Based Method}~\cite{gleixner_miplib_2021}: This method represents an instance in a 100-dimensional space by extracting features. The similarity between two instances is calculated as the Euclidean distance within this feature space. We denote this method as \emph{Feat}. The feature extraction code is publicly available~\footnote{\url{https://MIPlib.zib.de/downloads/feature_extractor.zip}}, and we utilized this implementation for all experiments. Notably, a direct comparison of this method to other baselines has not been conducted in previous works.
    \item \textbf{ISS}~\cite{steever_image-based_2022}: In this approach, instances are converted into grayscale images, which are subsequently fed into an autoencoder. The similarity between instances is computed as the Euclidean distance between the autoencoder's output vectors. We denote this method as \emph{ISS}. Since the authors did not provide an available implementation, we relied on the reported results from the original paper for the first part of our experiments. However, this method is excluded from the second part of our analysis due to the unavailability of an open-source implementation.%%%%%%%%%%%%PEUT GAGNER DE LA PLACE ICI
    \item \textbf{GNN Similarity}~\cite{steever_graph-based_2024}: This method leverages a GNN trained to classify instances into 19 predefined categories, using the reference set of 40 instances per class from the first experiment. The similarity between instances is defined as the Euclidean distance between the output embeddings of the GNN. We denote this method as \emph{GNN}. Although the official repository provides a pre-trained model, it does not include the necessary code for constructing or training the GNN. Consequently, we rely on the reported results from the corresponding paper for experiments under the same framework and utilize the pre-trained GNN model for additional experiments.
\end{itemize}

\subsection{Identifying Membership to Similar Classes}
%%%%%%%%%%%%%%%%%%%% PEUT GAGNER DE LA PLACE ICC
The primary objective of this evaluation is to assess the ability of our method to effectively group instances belonging to the same class. To ensure a fair comparison, we adopt the same experimental framework described in~\cite{steever_graph-based_2024}, leveraging their specified test and reference sets. This approach allows for direct performance comparisons with the \emph{ISS} and \emph{GNN} methods by using the data provided in~\cite{steever_graph-based_2024}. Additionally, we evaluate the performance of the \emph{Features} method.
The experiment considers 19 problem classes from the strIPLIB dataset. The abbreviations used for these classes are detailed in~\cite{steever_graph-based_2024,steever_image-based_2022}. These classes were justified in~\cite{steever_image-based_2022} as follows:  
\emph{"19 problem types which contained at least 50 total instances across all sources for that problem type"}. However, the specific selection process for these 50 instances per class is not detailed.
It is also noteworthy that, since the release of the library, 6 additional problem classes now contain more than 50 instances. Furthermore, some previously listed classes, such as \emph{bpp}, \emph{bp2}, and \emph{bif}, have since been reclassified as subclasses.

\begin{table}[htbp]
    \centering
    \setlength{\tabcolsep}{4.5pt}
    \begin{tabular}{c|c|c|c|c|c|c|c|c|c}
& \multicolumn{6}{c|}{Formal}& Feat&ISS&GNN\\
\hline
&$\cancel{\alpha}$&$\cancel{\beta}$&$\cancel{\gamma}$&$\cancel{\zeta}$&$\emptyset E$&$\emptyset$\\
bpp&\cellcolor[cmyk]{0.037,0.963,1.0,0.241} 42&\cellcolor[cmyk]{0.16,0.84,1.0,0.21} 47&\cellcolor[cmyk]{0.204,0.796,1.0,0.199} 49&\cellcolor[cmyk]{0.0,1.0,1,0.25} 26&\cellcolor[cmyk]{0.167,0.833,1.0,0.208} 47&\cellcolor[cmyk]{0.16,0.84,1.0,0.21} 47&\cellcolor[cmyk]{0.0,1.0,1,0.25} 27&\cellcolor[cmyk]{0.0,1.0,1,0.25} 32&\cellcolor[cmyk]{1.0,0.0,1,0.0} 81\\
bp2&\cellcolor[cmyk]{0.25,0.75,1.0,0.188} 62&\cellcolor[cmyk]{0.415,0.585,1.0,0.146} 71&\cellcolor[cmyk]{1.0,0.0,1,0.0} 100&\cellcolor[cmyk]{0.415,0.585,1.0,0.146} 71&\cellcolor[cmyk]{0.91,0.09,1.0,0.023} 95&\cellcolor[cmyk]{0.9,0.1,1.0,0.025} 95&\cellcolor[cmyk]{0.0,1.0,1,0.25} 24&\cellcolor[cmyk]{0.04,0.96,1.0,0.24} 52&\cellcolor[cmyk]{1.0,0.0,1,0.0} 100\\
bif&\cellcolor[cmyk]{1.0,0.0,1,0.0} 50&\cellcolor[cmyk]{0.76,0.24,1.0,0.06} 44&\cellcolor[cmyk]{1.0,0.0,1,0.0} 50&\cellcolor[cmyk]{1.0,0.0,1,0.0} 50&\cellcolor[cmyk]{1.0,0.0,1,0.0} 50&\cellcolor[cmyk]{1.0,0.0,1,0.0} 50&\cellcolor[cmyk]{0.0,1.0,1,0.25} 21&\cellcolor[cmyk]{0.28,0.72,1.0,0.18} 32&\cellcolor[cmyk]{0.72,0.28,1.0,0.07} 43\\
clp&\cellcolor[cmyk]{0.694,0.306,1.0,0.076} 54&\cellcolor[cmyk]{0.851,0.149,1.0,0.037} 59&\cellcolor[cmyk]{0.161,0.839,1.0,0.21} 37&\cellcolor[cmyk]{1.0,0.0,1,0.0} 64&\cellcolor[cmyk]{0.812,0.188,1.0,0.047} 57&\cellcolor[cmyk]{0.773,0.227,1.0,0.057} 56&\cellcolor[cmyk]{0.0,1.0,1,0.25} 28&\cellcolor[cmyk]{0.0,1.0,1,0.25} 12&\cellcolor[cmyk]{0.161,0.839,1.0,0.21} 37\\
col&\cellcolor[cmyk]{1.0,0.0,1,0.0} 40&\cellcolor[cmyk]{1.0,0.0,1,0.0} 40&\cellcolor[cmyk]{1.0,0.0,1,0.0} 40&\cellcolor[cmyk]{1.0,0.0,1,0.0} 40&\cellcolor[cmyk]{1.0,0.0,1,0.0} 40&\cellcolor[cmyk]{1.0,0.0,1,0.0} 40&\cellcolor[cmyk]{0.363,0.637,1.0,0.159} 27&\cellcolor[cmyk]{0.8,0.2,1.0,0.05} 36&\cellcolor[cmyk]{1.0,0.0,1,0.0} 40\\
cpm&\cellcolor[cmyk]{0.0,1.0,1,0.25} 49&\cellcolor[cmyk]{0.075,0.925,1.0,0.231} 54&\cellcolor[cmyk]{0.025,0.975,1.0,0.244} 51&\cellcolor[cmyk]{0.04,0.96,1.0,0.24} 52&\cellcolor[cmyk]{0.295,0.705,1.0,0.176} 65&\cellcolor[cmyk]{0.075,0.925,1.0,0.231} 54&\cellcolor[cmyk]{0.0,1.0,1,0.25} 32&\cellcolor[cmyk]{0.0,1.0,1,0.25} 47&\cellcolor[cmyk]{1.0,0.0,1,0.0} 100\\
cut&\cellcolor[cmyk]{0.0,1.0,1,0.25} 26&\cellcolor[cmyk]{0.0,1.0,1,0.25} 48&\cellcolor[cmyk]{0.182,0.818,1.0,0.205} 57&\cellcolor[cmyk]{0.0,1.0,1,0.25} 34&\cellcolor[cmyk]{0.0,1.0,1,0.25} 49&\cellcolor[cmyk]{0.0,1.0,1,0.25} 48&\cellcolor[cmyk]{0.0,1.0,1,0.25} 20&\cellcolor[cmyk]{0.0,1.0,1,0.25} 16&\cellcolor[cmyk]{1.0,0.0,1,0.0} 99\\
cvr&\cellcolor[cmyk]{0.753,0.247,1.0,0.062} 82&\cellcolor[cmyk]{0.344,0.656,1.0,0.164} 62&\cellcolor[cmyk]{0.866,0.134,1.0,0.034} 87&\cellcolor[cmyk]{0.968,0.032,1.0,0.008} 92&\cellcolor[cmyk]{0.962,0.038,1.0,0.009} 91&\cellcolor[cmyk]{0.962,0.038,1.0,0.009} 91&\cellcolor[cmyk]{0.0,1.0,1,0.25} 15&\cellcolor[cmyk]{0.0,1.0,1,0.25} 40&\cellcolor[cmyk]{1.0,0.0,1,0.0} 93\\
cwl&\cellcolor[cmyk]{1.0,0.0,1,0.0} 100&\cellcolor[cmyk]{0.95,0.05,1.0,0.013} 97&\cellcolor[cmyk]{0.9,0.1,1.0,0.025} 95&\cellcolor[cmyk]{0.95,0.05,1.0,0.013} 97&\cellcolor[cmyk]{1.0,0.0,1,0.0} 100&\cellcolor[cmyk]{0.95,0.05,1.0,0.013} 97&\cellcolor[cmyk]{1.0,0.0,1,0.0} 100&\cellcolor[cmyk]{0.0,1.0,1,0.25} 23&\cellcolor[cmyk]{1.0,0.0,1,0.0} 100\\
gap&\cellcolor[cmyk]{1.0,0.0,1,0.0} 100&\cellcolor[cmyk]{1.0,0.0,1,0.0} 100&\cellcolor[cmyk]{0.96,0.04,1.0,0.01} 98&\cellcolor[cmyk]{0.955,0.045,1.0,0.011} 98&\cellcolor[cmyk]{1.0,0.0,1,0.0} 100&\cellcolor[cmyk]{1.0,0.0,1,0.0} 100&\cellcolor[cmyk]{0.0,1.0,1,0.25} 22&\cellcolor[cmyk]{0.0,1.0,1,0.25} 30&\cellcolor[cmyk]{1.0,0.0,1,0.0} 100\\
inr&\cellcolor[cmyk]{0.265,0.735,1.0,0.184} 56&\cellcolor[cmyk]{0.396,0.604,1.0,0.151} 61&\cellcolor[cmyk]{0.259,0.741,1.0,0.185} 55&\cellcolor[cmyk]{0.926,0.074,1.0,0.019} 84&\cellcolor[cmyk]{0.442,0.558,1.0,0.14} 63&\cellcolor[cmyk]{0.442,0.558,1.0,0.14} 63&\cellcolor[cmyk]{1.0,0.0,1,0.0} 88&\cellcolor[cmyk]{0.0,1.0,1,0.25} 19&\cellcolor[cmyk]{0.915,0.085,1.0,0.021} 84\\
kps&\cellcolor[cmyk]{1.0,0.0,1,0.0} 100&\cellcolor[cmyk]{1.0,0.0,1,0.0} 100&\cellcolor[cmyk]{1.0,0.0,1,0.0} 100&\cellcolor[cmyk]{1.0,0.0,1,0.0} 100&\cellcolor[cmyk]{1.0,0.0,1,0.0} 100&\cellcolor[cmyk]{1.0,0.0,1,0.0} 100&\cellcolor[cmyk]{0.0,1.0,1,0.25} 35&\cellcolor[cmyk]{0.64,0.36,1.0,0.09} 82&\cellcolor[cmyk]{1.0,0.0,1,0.0} 100\\
lot&\cellcolor[cmyk]{0.115,0.885,1.0,0.221} 48&\cellcolor[cmyk]{0.328,0.672,1.0,0.168} 57&\cellcolor[cmyk]{0.661,0.339,1.0,0.085} 72&\cellcolor[cmyk]{0.132,0.868,1.0,0.217} 49&\cellcolor[cmyk]{0.552,0.448,1.0,0.112} 68&\cellcolor[cmyk]{0.322,0.678,1.0,0.17} 56&\cellcolor[cmyk]{0.0,1.0,1,0.25} 36&\cellcolor[cmyk]{0.0,1.0,1,0.25} 28&\cellcolor[cmyk]{1.0,0.0,1,0.0} 87\\
map&\cellcolor[cmyk]{0.139,0.861,1.0,0.215} 55&\cellcolor[cmyk]{0.33,0.67,1.0,0.168} 65&\cellcolor[cmyk]{0.376,0.624,1.0,0.156} 67&\cellcolor[cmyk]{0.0,1.0,1,0.25} 42&\cellcolor[cmyk]{0.278,0.722,1.0,0.18} 62&\cellcolor[cmyk]{0.33,0.67,1.0,0.168} 65&\cellcolor[cmyk]{0.0,1.0,1,0.25} 33&\cellcolor[cmyk]{0.0,1.0,1,0.25} 27&\cellcolor[cmyk]{1.0,0.0,1,0.0} 97\\
pcp&\cellcolor[cmyk]{0.19,0.81,1.0,0.203} 59&\cellcolor[cmyk]{0.735,0.265,1.0,0.066} 87&\cellcolor[cmyk]{0.825,0.175,1.0,0.044} 91&\cellcolor[cmyk]{0.735,0.265,1.0,0.066} 87&\cellcolor[cmyk]{0.67,0.33,1.0,0.083} 83&\cellcolor[cmyk]{0.735,0.265,1.0,0.066} 87&\cellcolor[cmyk]{0.0,1.0,1,0.25} 33&\cellcolor[cmyk]{0.16,0.84,1.0,0.21} 57&\cellcolor[cmyk]{1.0,0.0,1,0.0} 100\\
rel&\cellcolor[cmyk]{0.487,0.513,1.0,0.128} 28&\cellcolor[cmyk]{1.0,0.0,1,0.0} 38&\cellcolor[cmyk]{0.632,0.368,1.0,0.092} 31&\cellcolor[cmyk]{0.447,0.553,1.0,0.138} 27&\cellcolor[cmyk]{0.987,0.013,1.0,0.003} 38&\cellcolor[cmyk]{0.987,0.013,1.0,0.003} 38&\cellcolor[cmyk]{0.0,1.0,1,0.25} 18&\cellcolor[cmyk]{0.158,0.842,1.0,0.211} 22&\cellcolor[cmyk]{0.316,0.684,1.0,0.171} 25\\
sch&\cellcolor[cmyk]{1.0,0.0,1,0.0} 100&\cellcolor[cmyk]{1.0,0.0,1,0.0} 100&\cellcolor[cmyk]{1.0,0.0,1,0.0} 100&\cellcolor[cmyk]{0.99,0.01,1.0,0.003} 99&\cellcolor[cmyk]{1.0,0.0,1,0.0} 100&\cellcolor[cmyk]{1.0,0.0,1,0.0} 100&\cellcolor[cmyk]{0.0,1.0,1,0.25} 45&\cellcolor[cmyk]{0.0,1.0,1,0.25} 24&\cellcolor[cmyk]{0.74,0.26,1.0,0.065} 87\\
tup&\cellcolor[cmyk]{0.945,0.055,1.0,0.014} 97&\cellcolor[cmyk]{0.995,0.005,1.0,0.001} 100&\cellcolor[cmyk]{1.0,0.0,1,0.0} 100&\cellcolor[cmyk]{0.85,0.15,1.0,0.037} 93&\cellcolor[cmyk]{1.0,0.0,1,0.0} 100&\cellcolor[cmyk]{0.995,0.005,1.0,0.001} 100&\cellcolor[cmyk]{0.285,0.715,1.0,0.179} 64&\cellcolor[cmyk]{0.5,0.5,1.0,0.125} 75&\cellcolor[cmyk]{1.0,0.0,1,0.0} 100\\
vrp&\cellcolor[cmyk]{1.0,0.0,1,0.0} 100&\cellcolor[cmyk]{1.0,0.0,1,0.0} 100&\cellcolor[cmyk]{1.0,0.0,1,0.0} 100&\cellcolor[cmyk]{1.0,0.0,1,0.0} 100&\cellcolor[cmyk]{1.0,0.0,1,0.0} 100&\cellcolor[cmyk]{1.0,0.0,1,0.0} 100&\cellcolor[cmyk]{0.125,0.875,1.0,0.219} 56&\cellcolor[cmyk]{1.0,0.0,1,0.0} 100&\cellcolor[cmyk]{1.0,0.0,1,0.0} 100\\\hline\hline
mean&\cellcolor[cmyk]{0.589,0.411,1.0,0.103} 66&\cellcolor[cmyk]{0.692,0.308,1.0,0.077} 70&\cellcolor[cmyk]{0.757,0.243,1.0,0.061} 73&\cellcolor[cmyk]{0.66,0.34,1.0,0.085} 69&\cellcolor[cmyk]{0.792,0.208,1.0,0.052} 74&\cellcolor[cmyk]{0.766,0.234,1.0,0.059} 73&\cellcolor[cmyk]{0.0,1.0,1,0.25} 38&\cellcolor[cmyk]{0.0,1.0,1,0.25} 40&\cellcolor[cmyk]{1.0,0.0,1,0.0} 83

    \end{tabular}
    \caption{Evaluation of  similarity measures to identify whether instances belong to the same problem class.}
    \label{striplib_1}
\end{table}

The experimental results are summarized in Table~\ref{striplib_1}. Focusing first on the different variants of our method \emph{Formal}, we observe that removing any component of the distance (by setting its weight to zero) generally leads to a drop in performance. The full version ($\emptyset$) consistently outperforms both $\cancel{\alpha}$ and $\cancel{\zeta}$—the latter in all but one class—and is marginally superior to $\cancel{\beta}$ in all but three classes. The variant without right-hand side comparison ($\cancel{\gamma}$) yields results that are, on average, similar to the full version, but shows a significantly higher worst-case drop (up to 19\%), indicating greater variability and instability. When comparing the greedy version ($\emptyset$) to the exact formulation ($\emptyset E$), results are identical in 17 out of 19 classes. The average runtime to compute the top-40 neighbors per test instance is $2.8\times 10^{-3}$ seconds (std $4.4\times 10^{-3}$) for the greedy version, compared to $0.37$ seconds (std $0.65$) for the exact method. This highlights not only a 200$\times$ speedup but also a much more stable runtime, making the greedy variant clearly preferable in practice.
In comparison to the baselines, both our method (\emph{Formal}) and the supervised \emph{GNN} approach consistently outperform \emph{Features} and \emph{ISS}. The latter two perform at least $50\%$ worse than the best score in 12 out of the 19 evaluated classes. While \emph{GNN} benefits from supervised training on nearly 800 labeled instances from similar classes, the performance gap between the two remains small. \emph{Formal} achieves results within $5\%$ of the best score in 12 out of 19 classes, compared to 14 out of 19 for \emph{GNN}, with similarly close average accuracy.

Regarding class-wise performance, more than ten classes exhibit near-perfect top-40 accuracy above $95\%$, while others perform significantly worse. Further analysis (not shown) highlights notable structural overlaps between certain classes. For instance, at least $10\%$ of the top 40 most similar instances retrieved by \emph{Formal} in some cases belong to a different class, and vice versa. Notably, \emph{bpp} and \emph{cut} are explicitly marked as structurally similar in the StrIPLIB documentation. Likewise, \emph{cpm} and \emph{cwl} both fall under capacity-constrained formulations, and \emph{map} and \emph{col} share similar structures despite being distinct classes. The \emph{rel} class also performs poorly across all metrics, likely due to its high internal heterogeneity and overlap with the \emph{bif} class. These overlaps appear to result from the inherently ambiguous nature of instance class definitions, which often blur structural boundaries between problems.

\subsection{Identifying Belongings to  Similar Subclasses}

\begin{table}[htbp]
    \centering
    \begin{tabular}{c|m{2.8cm}|c|c|c}
&& Formal & Feat& GNN\\\hline
\multirow{4}{*}{bpp} & conflicts&\cellcolor[cmyk]{1.0,0.0,1,0.0} 0.51&\cellcolor[cmyk]{0.2709359605911332,0.7290640394088668,1.0,0.1822660098522167} 0.32&\cellcolor[cmyk]{0.9802955665024635,0.019704433497536478,1.0,0.004926108374384119} 0.5\\
 & item-fragmentation&\cellcolor[cmyk]{0.03661971830985913,0.9633802816901409,1.0,0.24084507042253522} 0.46&\cellcolor[cmyk]{0.0,1.0,1,0.25} 0.37&\cellcolor[cmyk]{1.0,0.0,1,0.0} 0.89\\
 & plain&\cellcolor[cmyk]{0.9849999999999999,0.015000000000000124,1.0,0.003750000000000031} 0.99&\cellcolor[cmyk]{0.0,1.0,1,0.25} 0.28&\cellcolor[cmyk]{1.0,0.0,1,0.0} 1.0\\
 & two-dimensional&\cellcolor[cmyk]{0.6849999999999998,0.31500000000000017,1.0,0.07875000000000004} 0.84&\cellcolor[cmyk]{0.0,1.0,1,0.25} 0.28&\cellcolor[cmyk]{1.0,0.0,1,0.0} 1.0\\\hline
\multirow{4}{*}{lot} & linked-lot-sizes&\cellcolor[cmyk]{1.0,0.0,1,0.0} 1.0&\cellcolor[cmyk]{0.0,1.0,1,0.25} 0.31&\cellcolor[cmyk]{0.0,1.0,1,0.25} 0.48\\
 &sizing multi-level&\cellcolor[cmyk]{0.7095435684647302,0.2904564315352698,1.0,0.07261410788381745} 0.52&\cellcolor[cmyk]{0.5684647302904563,0.4315352697095437,1.0,0.10788381742738593} 0.47&\cellcolor[cmyk]{1.0,0.0,1,0.0} 0.6\\
 &sizing multi-level-with-setup-time&\cellcolor[cmyk]{0.8342541436464092,0.16574585635359085,1.0,0.04143646408839771} 0.42&\cellcolor[cmyk]{0.4475138121546963,0.5524861878453037,1.0,0.13812154696132592} 0.33&\cellcolor[cmyk]{1.0,0.0,1,0.0} 0.45\\
 &sizing single-level&\cellcolor[cmyk]{1.0,0.0,1,0.0} 1.0&\cellcolor[cmyk]{0.0,1.0,1,0.25} 0.47&\cellcolor[cmyk]{0.0,1.0,1,0.25} 0.46\\\hline
\multirow{3}{*}{vrp} & capacitated&\cellcolor[cmyk]{0.9249999999999998,0.07500000000000018,1.0,0.018750000000000044} 0.96&\cellcolor[cmyk]{0.0,1.0,1,0.25} 0.21&\cellcolor[cmyk]{1.0,0.0,1,0.0} 1.0\\
& concrete-delivery&\cellcolor[cmyk]{0.46365914786967405,0.5363408521303259,1.0,0.13408521303258147} 0.73&\cellcolor[cmyk]{0.0025062656641603475,0.9974937343358397,1.0,0.2493734335839599} 0.5&\cellcolor[cmyk]{1.0,0.0,1,0.0} 1.0\\
 & time-windows&\cellcolor[cmyk]{1.0,0.0,1,0.0} 1.0&\cellcolor[cmyk]{0.0,1.0,1,0.25} 0.38&\cellcolor[cmyk]{1.0,0.0,1,0.0} 1.0\\\hline\hline
 \multicolumn{2}{c}{mean}&\cellcolor[cmyk]{1.0,0.0,1,0.0} 77&\cellcolor[cmyk]{0.0,1.0,1,0.25} 36&\cellcolor[cmyk]{0.99,0.01,1.0,0.003} 76
    \end{tabular}
    \caption{Evaluation of similarity measures to identify whether instances belong to the same subclass.}
    \label{striplib_2}
\end{table}

The objective of this second part of the simulation is analogous to the first; however, it focuses on problem subclasses, making it intrinsically more complex as it requires distinguishing between sets of more closely related instances. 
To maintain consistency within the training framework of the \emph{GNN} method and to ensure a fair comparison, the subclasses studied belong exclusively to the classes trained by \emph{GNN}.
We select the set of subclasses based on the following criteria: (i) the class must be part of the \emph{GNN} reference set; (ii) the subclass must contain at least $50$ instances; and (iii) there must be more than one eligible subclass per class. This selection yields three classes, each with three to four subclasses. Notably, three of the four subclasses within the \emph{bpp} class were explicitly trained as separate classes by the \emph{GNN}, potentially giving it a specific advantage. The instances are selected using the seed $\{1\}$ to ensure reproducibility.

The results of this experiment are presented in Table~\ref{striplib_2}. Once again, the \emph{Features} method performs significantly worse compared to both \emph{Formal} and \emph{GNN}. However, it remains challenging to definitively determine superiority between \emph{Formal} and \emph{GNN}. For the \emph{bpp} class, \emph{GNN} outperforms \emph{Formal} by $48\%$ and $16\%$ on two of the subclasses, respectively, with both methods achieving similar results on the remaining two subclasses, although \emph{GNN} was specifically trained to recognize these subclasses. For the \emph{lot} class, each method excels in two of the four subclasses. However, while \emph{GNN} underperforms by at least $52\%$, \emph{Formal}'s performance is never more than $6\%$ below the best, demonstrating greater reliability across these subclasses. For the \emph{vrp} class, \emph{GNN} achieves perfect classification, while \emph{Formal} closely follows, with an average gap of approximately $10\%$. Overall, both methods reach comparable mean performance, with \emph{Formal} slightly ahead by $1\%$.

To summarize these experiments, \emph{Formal} significantly outperforms all unsupervised baselines (\emph{ISS} and \emph{Features}). When compared to \emph{GNN}, neither method clearly surpasses the other. However, \emph{GNN} benefits from supervised training on this specific set of problem classes using a labeled dataset of 760 instances. These problem classes have inherent limitations, and \emph{GNN} is confined to its training framework, making the generalizability of its similarity metric across the entire MILP space impractical.

\section{Conclusion and Future Works}
In this paper, we introduced a novel similarity metric for MILP that exhibits the mathematical properties of a distance function. This metric is derived solely from the intrinsic data of the problems being compared, without requiring any training process. We demonstrated that our proposed metric outperforms the considered baselines, particularly challenging a classification method that relies on a labeled training dataset.

Our proposed similarity metric offers several advantages. First, it can effectively characterize the heterogeneity within a set of instances, making it applicable to various use cases, such as selecting heterogeneous benchmark instances (e.g., MIPLIB) or quantifying the diversity of any evaluation set. Second, it defines instance classes based solely on the structural representation of constraints, without needing to consider the specific objective of the instances. This approach not only bridges existing instance classes but also enables the association of unlabeled or sparsely labeled instances with known groups.

Future work will focus on extending this methodology to develop a classification of the entire MILP space, with the aim of generalizing ML-MILP methods across all MILP instances. As highlighted in previous studies, ML methods for MILP solvers (e.g., branching, searching, cutting) typically perform well only on structurally similar instances. A comprehensive classification of the MILP space will facilitate the development of specialized ML-MILP models targeting specific instance classes, improving their overall effectiveness.

\section*{Acknowledgements}  
This research was supported by the Agence Nationale de la Recherche (grant ANR-22-CE46-0011) and the Luxembourg National Research Fund (grant INTER/ANR/22/17133848) through the UltraBO Project. Computational experiments were carried out using the High-Performance Computing (HPC) facilities of the University of Luxembourg.

We also thank Mark Turner for insightful discussions, Pierre Talbot for his unwavering support, and Louis Gasse for his assistance with the mathematical formulation of the problem.

\bibliography{references}

\end{document}